\documentclass{article}

\usepackage{arxiv}
\renewcommand{\shorttitle}{GraphWalk: Enabling Reasoning in LLMs through Tool-Based Graph Navigation}
\usepackage[utf8]{inputenc} 
\usepackage[T1]{fontenc}    
\usepackage{hyperref}       
\usepackage{url}            
\usepackage{booktabs}       
\usepackage{amsfonts}       
\usepackage{nicefrac}       
\usepackage{microtype}      
\usepackage{lipsum}		
\usepackage{graphicx}
\usepackage{natbib}
\usepackage{doi}
\usepackage{multirow} 
\usepackage{inconsolata}
\usepackage{times}
\usepackage{latexsym}
\usepackage{listings}
\usepackage{longtable} 
\usepackage{array}
\usepackage{amsmath}
\usepackage{placeins}

\title{GraphWalk: Enabling Reasoning in Large Language Models through Tool-Based Graph Navigation}

\author{{Taraneh Ghandi}\\
	Faculty of Science\\
	McMaster University\\
	\texttt{ghandit@mcmaster.ca} \\
	\And
	{Hamidreza Mahyar} \\
	Faculty of Science\\
	McMaster University\\
	\texttt{mahyarh@mcmaster.ca} \\
	\AND
	{Shachar Klaiman} \\
	BASF Digital Solutions \\
	\texttt{shachar.klaiman@basf.com} 
}

\hypersetup{
pdftitle={GraphWalk-EnablingReasoningInLargeLanguageModelsThroughToolBasedGraphNavigation},
pdfsubject={q-bio.NC, q-bio.QM},
pdfauthor={David S.~Hippocampus, Elias D.~Striatum},
pdfkeywords={LLM/AI Agents, Knowledge Graphs, Multihop QA, Reasoning, Graph-Based Methods, Knowledge Tracing, Benchmarking, Evaluation Methodologies},
}

\begin{document}
\maketitle

\begin{abstract}
    The use of knowledge graphs (KG) for grounding agents in real-world Q\&A applications has become increasingly common. In these applications, answering complex queries often requires multi-hop reasoning and as well as the ability to navigate through the vast relational structures. Standard approaches rely on different prompting techniques that steer the large language models (LLMs) to reason over the raw graph context or retrieval-augmented generation pipelines where relevant subgraphs are injected into the context. These, however, face severe limitation when facing enterprise-scale KGs which cannot fit in even the largest context-windows available today. Here, we present \textbf{\textit{GraphWalk}}, a problem-agnostic, training-free, and tool-based framework that allows off-the-shelf LLMs to reason through sequential graph navigation in order to dramatically increase the performance over different tasks. Unlike task-specific agent frameworks that encode domain knowledge into specialized tools, GraphWalk equips the LLM with a minimal set of orthogonal graph operations just enough to traverse any graph structure. We then evaluate whether models equipped by GraphWalk can compose these operations into correct multi-step reasoning chains. Each tool call represents a verifiable reasoning step, creating a transparent execution trace. We first demonstrate our approach on a small maze traversal problem, a problem which non-reasoning models are completely unable to solve. We then present our solution for graph structures that more closely resemble real-world enterprise knowledge graphs. To isolate structural reasoning from world knowledge, we evaluate on entirely synthetic graphs with random, non-semantic labels. Our benchmark spans 12 query templates from basic retrieval to compound first-order logic queries. Results show that tool-based traversal yields substantial and consistent performance gains over in-context baselines across all model families tested, and that these gains become more pronounced as problem scale increases, precisely where in-context approaches fail catastrophically.
\end{abstract}


\section{Introduction}

The introduction of reasoning capabilities into large language models (LLMs) has dramatically advanced the ability of LLMs to solve complex tasks in different domains \cite{wei2022chain, hendrycks2021measuring, cobbe2021training, chen2021evaluating}. The ability to reason, however, still critically depends on having the relevant information available in the model’s context window, specially for domain-specific, private or frequently changing data. Even with ever-increasing context window sizes, many real world applications require reasoning over substantial amount of data which would likely never fit into the context window.
One example is a supply chain analysis application for a large enterprise. The supply chain network in such companies can be very extensive, often starting from hundreds of thousands of nodes and relationships. In order to answer user queries and run different analysis over the data, the LLM agent needs to be able to go through the graph structure, gather the relevant information, analyze it, and compose a final answer. This remains a challenge even with the existing reasoning models and large context windows.
There have been many efforts attempting to help LLMs perform better on such tasks, from specialized agents \cite{jiang2025kg}, to pre-trained subgraph retrieval models \cite{zhang2022subgraph, zhang2022greaselm, he2024gretriever}, query code generation methods \cite{feng2024cypherbench, d2025investigating, steinigen2024fact, zhong2024synthet2c} and others \cite{wang2025hierarchical, dao2025alphamaze}. While advancing the performance over different tasks, these approaches are often specialized to specific domains or require additional fine-tuning.

In attempt to go beyond these approaches, we aim to assess the fundamental capacity of LLMs for \textit{structural graph reasoning}. We first show how non-reasoning off-the-shelf LLMs completely fail to solve even a simple 10 $\times$ 10 maze traversal problem which small reasoning models are capable to resolve. We then show how representing the maze as a graph and augmenting the non-reasoning models with only a couple of simple navigation tools enables them to go from 0\% to 100\% permformance on this task. This suggests that if we can map a task onto a graph structure and equip the model with graph-navigation tools, even non-reasoning models can reason through the underlying information structure and solve complex tasks.
We then demonstrate our approach for more real-world, enterpirse resembling graphs.
Two main elements contribute to this approach. The first element consists of a \textit{minimal}, problem-agnostic toolset containing only basic graph navigation primitives: node lookup, neighbor retrieval, and property enumeration. The agent must compose these tools to solve complex queries, and the tools themselves carry no domain intelligence. The second element is the evaluation of \textit{entirely synthetic graphs} with random and non-semantic labels, with the intent to eliminate any possibility of parametric knowledge leakage. Together, these elements ensure that any observed performance reflects genuine structural reasoning rather than memorized associations or tool-encoded shortcuts.

This framing yields the \textbf{\textit{GraphWalk}} framework, which positions the LLM as an agent that navigates a graph by executing tool calls sequentially to gather information for answering questions. Each tool call represents a verifiable reasoning step, producing a transparent execution trace.

Our contributions are:
\begin{enumerate}
    \item \textit{A Problem-Agnostic Framework for Autonomous Graph Traversal:} We introduce a tool-based agentic framework with a minimal, orthogonal toolset enabling LLMs to reason over arbitrarily large graph structures through sequential exploration, having only the graph schema and the given tools. Unlike task-specific agent designs, our tools carry no domain knowledge, focusing on the evaluation of the model's reasoning capacity.
    \item \textit{Maze Traversal as a Visual Demonstration:} We first validate GraphWalk on a small example of the maze traversal problem and show that tool-equipped agents can solve spatial reasoning tasks that LLMs are unable to solve when given simply in their context.
    \item \textit{Controlled Synthetic Graph Benchmark:} We design a configurable synthetic graph generator that produces entirely random graphs with non-semantic labels, along with a benchmark of 12 query categories spanning retrieval, aggregation, multi-hop traversal, and first-order logic operations, to evaluate pure structural reasoning capabilities.
\end{enumerate}

\section{Related Work}
\label{sec:related_work}

The effort to synergize Large Lanugage Models (LLMs) and Knowledge Graphs (KGs) has resulted in several distinct approaches:

    \paragraph{Graph Textualization and RAG:} These approaches inject graph structures into text through triple linearization or JSON representations for injection into LLM context windows. Benchmarks such as GrailQA \cite{gu2021beyond}, GrailQA++ \cite{dutt2023grailqapp}, and WebQSP \cite{yih2016value} evaluate question-answering over KG contexts. These Retrieval-Augmented Generation methods focus on retrieving and formatting relevant subgraphs \cite{edge2024local, zhang2022greaselm, zhang2022subgraph}, but struggle with scalability and preserving global graph structure.
    
    \paragraph{LLMs as Query Generators:} These approaches train or prompt LLMs to translate natural language questions into formal query languages like SPARQL or Cypher \cite{kovriguina2023sparqlgen, zahera2024generating, dabramo2025investigating, ozsoy2025text2cypher}. While benefiting from structured query engines, they frame tasks as one-shot translation problems, which are unsuitable for exploratory reasoning as query errors often result in correction loops without resolution.

    \paragraph{Hybrid LLM-GNN Architectures:} Methods integrating Graph Neural Networks with LLMs \cite{mavromatis2024gnn, liu2025dual} mostly use GNNs to learn topology-aware node and edge embeddings to provide compressed structural representations. This enriches LLM understanding through improved internal representations rather than explicit, verifiable tool-use behavior.

    \paragraph{Tool-Augmented Graph Agents:} Our work aligns with the emerging "LLM as Agent" paradigm applied to graphs. Frameworks like \texttt{Reason-Align-Respond} (RAR) \cite{shen2025reason} and \texttt{SubgraphRAG} \cite{li2025simple} demonstrate potential for working through complex queries through paradigms where LLMs actively reflect and take actions rather than passively read. KG-Agent \cite{jiang2025kg} is an example that presents a tool-based approach in which a 7B LLM is fined-tuned on code-based instruction data synthesized from Knowledge Graph Question Answering (KGQA) datasets and given 13 specialized tools. 

    \paragraph{Positioning \textit{GraphWalk}:} Unlike existing tool-augmented KG reasoning frameworks such as KG-Agent, GraphWalk assess the \textit{structural graph reasoning} capability of LLMs using only minimal, generic graph operations. Our toolset consists of just 2 tools for the maze traversal example and 4 tools for the random graph experiments,
    none of which encode domain-specific logic. By taking away task-specific scaffolding and semantic knowledge (via synthetic graphs), we isolate the reasoning component that all graph-LLM integration approaches ultimately depend on. 

\section{Maze Traversal as a Graph Reasoning Task}
\label{sec:maze}

\begin{figure}[t]
\centering
\includegraphics[width=0.4\columnwidth]{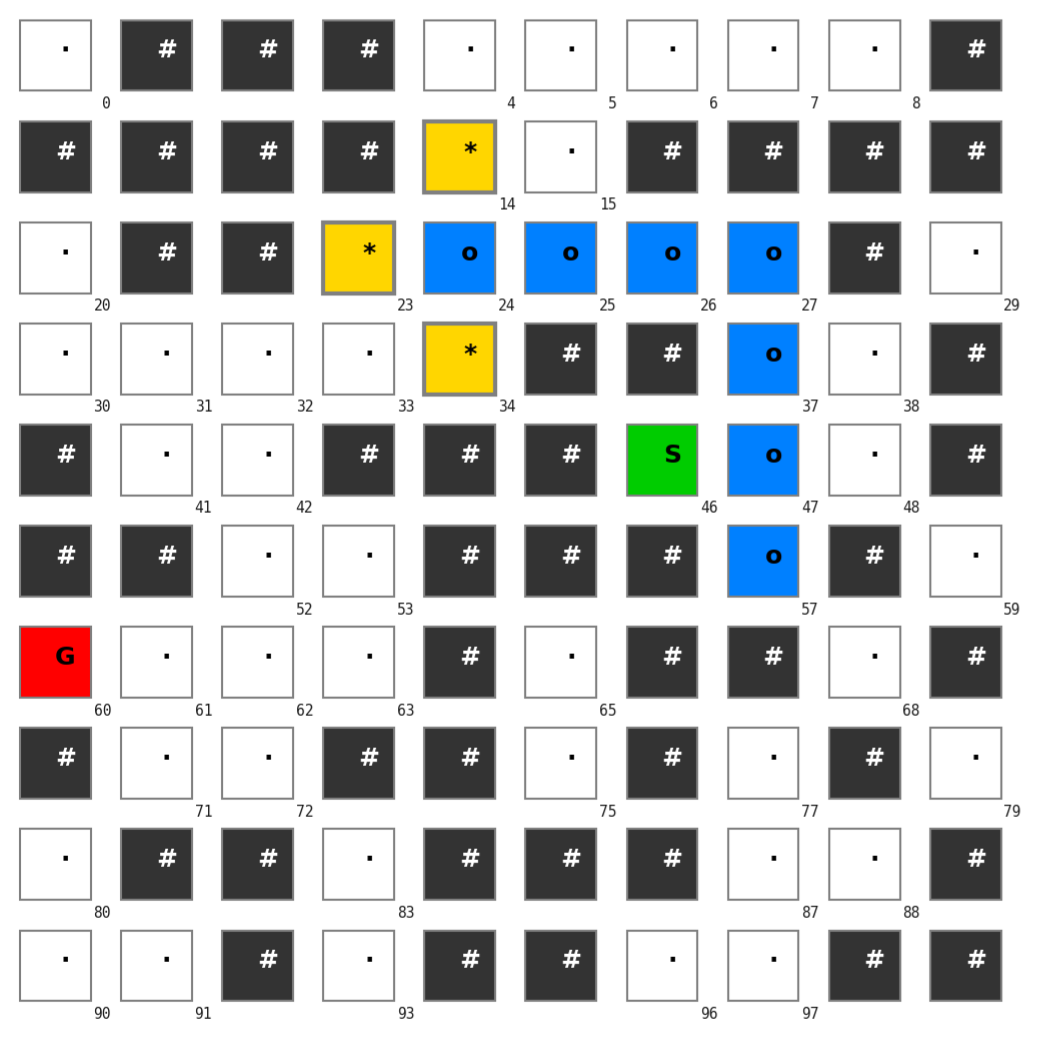} 
\caption{Agent exploration of a 10x10 maze mid-traversal at node 24, after calling \texttt{get\_next\_possible\_cells}. Yellow cells (*): tool output, blue (o): visited nodes, black (\#): walls, white (.): unvisited path cells, green (s): start, and red (G): goal.}
\label{fig:maze_example}
\end{figure}

In order to build intuition for GraphWalk, we first showcase its abilities on maze traversal, using this problem as a domain where the graph structure is clearly visible. This example illustrates the core mechanics of the framework before we move to graphs which resemble real-world enterprise KGs more, and thus less directly interpretable. 
Maze traversal requires capabilities that are fundamental to graph reasoning more broadly, as the solver must maintain state across decisions and reason about connectivity and spatial direction. Prior work has demonstrated that language models achieve near-zero accuracy on textual maze representations, even when the full maze fits within the model's context window \cite{wang2025hierarchical}.
\paragraph{Maze-as-Graph Representation:} We model a maze instance as a property graph stored in Neo4j \cite{neo4j}. Every cell in the maze grid is treated as a node in this graph, and \texttt{ADJACENT} relationships connect cells that share an edge (up, down, left, right). The mazes can be procedurally and randomly generated with configurable parameters: grid-size, wall ratio, and minimum path length between start and goal. The generation process guarantees that at least one valid path exists. To present a small example, we conduct our experiments on mazes of size 10 $\times$ 10. A figure displaying an agent's exploration of such maze is shown in \ref{fig:maze_example}.

\subsection{Traversal Toolset and Node Properties}

Consistent with GraphWalk's minimalist design philosophy, we equip the agent with two traversal tools and the graph schema. If at any point these tools result in error, an informative response is presented to the agent:

\paragraph{get\_possible\_next\_cells(node\_id):} Identifies all traversable cells directly adjacent (up, down, left, right) to the specified cell. The tool first verifies that the queried cell is itself a valid path cell (not a wall), then marks it as visited in the maze, and returns the list of neighboring cells that are traversable. At each step, this tool reveals which directions are open from the current position, enabling the agent to make decisions based on the node properties.

\paragraph{get\_connected\_path():} Retrieves all visited nodes and computes the connected path from the first to the last visited cell, requiring each step to be adjacent (no jumps or gaps). This tool allows the agent to validate the path taken so far. Curcially, this tool does not compute arbitrary shortest paths in the maze. It operates exclusively on visited nodes, preserving the constraint that knowledge must be earned through exploration. This tool is typically called once by the agent upon reaching the goal.

These tools mirror GraphWalk's core philosophy: they provide basic traversal operations without encoding any pathfinding intelligence. The agent must decide which cells to explore, when to backtrack, and when to compose its visited cells into a valid path. The cells in the graph carries the following properties:

\begin{itemize}
    \item \texttt{key}: Represents the key of the maze cell, acting as an identifier.
    \item \texttt{euclidean\_distance}: The Euclidean distance from the cell to the goal node. This serves as a directional heuristic and does not indicate actual path length, as walls may intervene. Wall nodes are assigned a sentinel value of $-10^9$ to signal inaccessibility.
    \item \texttt{marked}: Path cells contain this boolean flag, initially \texttt{False}, that is set to \texttt{True} when the agent visits a cell by calling the \texttt{get\_possible\_next\_cells} tool. This enables the agent to track its exploration history. 
    \item \texttt{mark\_order}: Path cells have an integer property recording the visitation order, initially set to $-1$.
\end{itemize}

\FloatBarrier

\begin{table}[t]
    \small
\centering
\vspace{0pt}
\setlength{\tabcolsep}{5pt}  
\begin{tabular}{llccccccc}
\hline
\textbf{Model} & \textbf{Tools} & \textbf{Correct} & \textbf{Accuracy} & \textbf{Avg IT} & \textbf{Avg TC} \\
\hline
\multirow{2}{*}{gpt-4o-mini} & True & 8/10 & 80.00 & 144.971 & 22.125\\
                             & False & 0/10 & 0.00 & 27.541 & -\\
\hline
\multirow{2}{*}{gpt-4o} & True & 6/10 & 60.00 & 135.196 & 15.1\\
                        & False & 1/10 & 10.00 & 14.159 & -\\
\hline
\multirow{2}{*}{gpt-4.1} & True & \textbf{10/10} & \textbf{100.00} & \textbf{127.299} & 21.4\\
                         & False & 1/10 & 10.00 & 13.060 & -\\
\hline
\multirow{2}{*}{gpt-4.1-mini} & True & \textbf{10/10} & \textbf{100.00} & 513.433 & 26.5\\
                              & False & 0/10 & 0.00 & 72.22 & -\\
\hline
\multirow{2}{*}{gpt-4.1-nano} & True & 0/10 & 0.00 & 277.074 & 16.2\\
                              & False & 0/10 & 0.00 & 13.013 & - \\
\hline
o3-mini & False & \textbf{10/10} & \textbf{100.00} & 46.548 & -\\
o4-mini & False & \textbf{10/10} & \textbf{100.00} & 43.007 & -\\
\hline
\end{tabular}
    \caption{Performance of LLM agents in maze traversal (10 mazes × 10 runs). "Tools" = tool use; "Avg IT" = average inference time (s); "Avg TC" = average tool calls. Reasoning models run without tools as a baseline. Best results for non-reasoning and reasoning models in boldface.}
    \label{tab:maze_results_single}
\end{table}

\subsection{Experiments, Evaluation Details and Results for Maze Traversal}

\paragraph{Experiments and Evaluation Details: }We use seven models: gpt-4.1-nano, gpt-4.1-mini, gpt-4.1 \cite{openai2025gpt41}, gpt-4o-mini, gpt-4o \cite{openai2024gpt4o}, o3-mini, and o4-mini \cite{openai2025o3o4} (with default reasoning effort for reasoning models). The generated mazes are 10 $\times$ 10 grids with 50\% of the maze cells randomly selected as wall cells, and with ground-truth paths of at least 15 steps. Each model is tested on 10 generated instances and the results are averaged. We evaluate whether the agent successfully finds \textit{any} valid path from the start to the goal node since the agent explores the maze incrementally through local observations and does not have a global view of the maze structure. This is a more difficult task than traditional maze traversal solutions. Requiring optimal solutions would be illogical since the agent, by design, discovers the graph through sequential tool calls. As a baseline comparison, we also evaluate models in a no-tool setting where the full maze is inserted in the LLM's context window, similar to traditional solutions for this problem and the previous works.

\paragraph{Maze Traversal Results}The maze traversal results are presented in \ref{tab:maze_results_single}. These results reinforce GraphWalk's central idea: in the no-tool setting, where models receive the complete maze fed into the context, they fail to find even valid paths, consistent with prior findings that LLMs achieve near-zero accuracy on maze tasks \cite{wang2025hierarchical}. In contrast, tool-equipped agents successfully find valid paths by decomposing the task into a sequence of local observations and movement decisions. The Euclidean distance heuristic provided as a node property plays an important role: it gives the agent a sense of directionality when choosing among neighboring cells, enabling it to make informed (though not necessarily optimal) decisions about which cells to explore next. This mirrors how the knowledge graph tools provide the agent with local structural information at each step, from which it must compose a global reasoning chain.
It is important to note that all mazes in our evaluation are 10 $\times$ 10 grids, small enough to fit entirely within any model's context window with substantial capacity to spare. This choice for the maze sizes is deliberate since it demonstrates how even in these small instances, the models fail to find valid paths without tool access and often produce answers that contain jumps between non-adjacent cells or wall nodes.

\section{Reasoning by Navigation on Synthetic Graphs}
\label{sec:methodology}
We now turn to the primary experimental setting of reasoning over property graphs that resemble real-world enterprise knowledge graphs more closely. These graphs are \textit{entirely random}, and in contrast to benchmarks based on factual domains like Freebase \cite{bollacker2008freebase} or Wikidata \cite{vrandevcic2014wikidata}, these graphs are synthetically generated with labels void of semantic meaning for nodes, properties, and relationships. The structure of these graphs is more relational than geometric, and unlike the graphs built for the maze traversal problem, they do not contain spatial information. The queries built for these graphs are more complex as compared to the agent merely finding its way to the goal and presenting a valid path. These assess the models' capability to answer questions from retrieval, aggregation, relational traversal, and logical composition categories.  

\subsection{Isolating Structural Reasoning from Parametric Knowledge}

Existing real-world KGQA benchmarks based on Freebase or Wikidata are problematic precisely because the knowledge they contain overlaps substantially with LLM training corpora. Prior work has shown that LLMs achieve more than 30\% accuracy on these benchmarks through prompting alone, without any access to the underlying graph \cite{jiang2025kg}. This indicates that a significant portion of questions can be answered from parametric knowledge, which confounds the evaluation of structural reasoning. Our synthetic graphs eliminate this confound entirely, ensuring that every correct answer must be derived through actual graph exploration and isolates the \textit{structural reasoning component} that underlies all graph-LLM integration approaches.

\subsection{Comparison to Existing Baseline Approaches}
\label{sec:baseline_relationship}

GraphWalk can be situated relative to subgraph extraction, GraphRAG, and text-to-Cypher approaches. Subgraph extraction and GraphRAG methods retrieve a relevant subgraph and then reason over it. These subgraphs are typically small, comparable in scale to the 100–500 node graphs in our experiments. Our no-tool baseline, which feeds the full graph into the context window, is functionally equivalent to a \textit{perfectly retrieved} subgraph with zero retrieval error. Yet models still struggle with reasoning on these small graphs (Table~\ref{tab:main_results_single}), suggesting that the core challenge lies in the reasoning step itself, not retrieval quality. Text-to-Cypher/SPARQL approaches treat KGQA as one-shot translation to a formal query. This is effective for well-structured questions but brittle for exploratory reasoning where the solution path is unknown. Query generation errors are difficult to recover from and often lead to unproductive correction loops (Section~\ref{sec:related_work}). GraphWalk's iterative exploration avoids this by allowing the agent to incrementally gather information and adjust its strategy based on intermediate results.

\subsection{The Agent's Toolset and Reasoning Loop}

The LLM agent is provided with a minimal yet sufficient set of tools for graph traversal:

     \paragraph{get\_node\_by\_property:} Retrieves nodes with a specified label that match a given property-value pair. This is the primary entry-point tool for locating specific entities in the graph when their identifying property (e.g., name, ID, or unique attribute) is known.

    \paragraph{get\_all\_nearest\_neighbors:} Returns all nodes directly connected to a specified node through any relationship type. This tool enables the exploration of a node's immediate neighborhood and the features of its direct relations.

    \paragraph{get\_unique\_property\_values:} Retrieves all distinct values for a specified property across nodes or relationships of a given type. This tool assists the agent in finding available entities and validating specific values before searching, which serves as a critical data exploration mechanism when the agent needs to enumerate possible options available for a particular property.

    \paragraph{think:} Records the agent's intermediate steps by taking a string input and returning it unchanged, creating a visible record in the agent's execution trace. This tool enables the agent to document its thought process, plan, and articulate decisions. 

\begin{figure}[t]
\centering
\includegraphics[width=0.6\columnwidth]{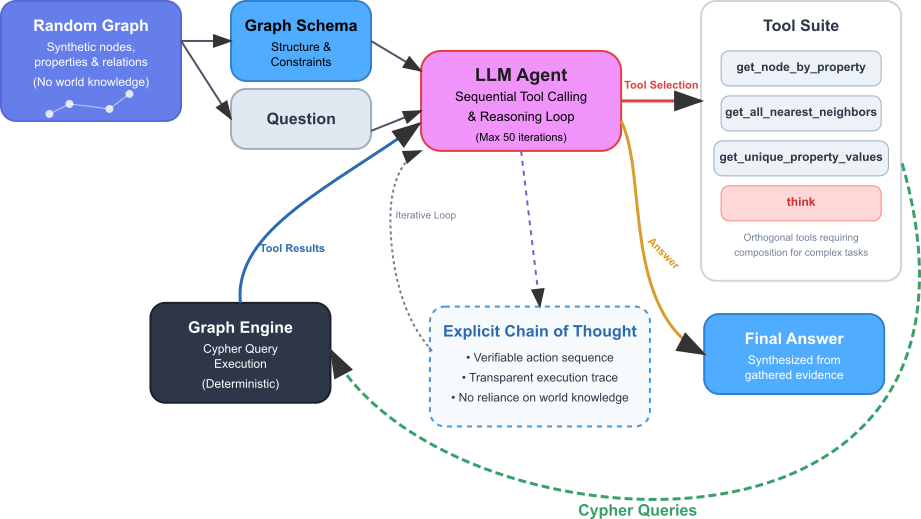} 
\caption{An overview of the agentic framework.}
\label{fig:framework_overview}
\end{figure}

Each tool is equivalent to a cypher query which is executed on the graph database, except for the \textit{\texttt{think}} tool. 
Given a question, the LLM selects tool(s) to \textit{act}, and then \textit{observes} the result returned from a deterministic graph executor. This cycle repeats until the agent decides that it has gathered sufficient information to formulate a final answer, or a limit of 30 iterations is reached. Figure \ref{fig:framework_overview} demonstrates an overview of our framework using these synthetic graphs. More details regarding the tools are laid out in appendix~\ref{appendix:agent_tools}.

\FloatBarrier

\begin{table}[t]
    \small
    \centering
    \vspace{0pt}
    \setlength{\tabcolsep}{3pt}  
    \begin{tabular}{llccccccc}
    \hline
    \textbf{Model} & \textbf{T} & \textbf{C} & \textbf{A} & \textbf{P} & \textbf{R} & \textbf{F1} & \textbf{FP} & \textbf{TC} \\
    \hline
    \multirow{2}{*}{gpt-4o-mini} & True & 18 & 15.00 & 0.23 & 0.25 & 0.23 & 170 & 7362\\
                                 & False & 6 & 5.00 & 0.35 & 0.24 & 0.26 & 350 & -\\
    \hline
    \multirow{2}{*}{gpt-4o} & True & 34 & 28.33 & 0.42 & 0.39 & 0.39 & 235 & 2548\\
                            & False & 18 & 15.00 & 0.50 & 0.44 & 0.44 & 300 & -\\
    \hline
    \multirow{2}{*}{gpt-4.1} & True & \textbf{35} & 29.17 & 0.47 & 0.41 & 0.43 & 232 & 2483\\
                             & False & 21 & 17.50 & 0.52 & 0.49 & 0.48 & 1006 & -\\
    \hline
    \multirow{2}{*}{gpt-4.1-mini} & True & 32 & 26.67 & 0.49 & 0.43 & 0.43 & 226 & 3561\\
                                  & False & 20 & 17.7 & 0.49 & 0.48 & 0.46 & 522 & -\\
    \hline
    \multirow{2}{*}{gpt-4.1-nano} & True & 11 & 9.17 & 0.16 & 0.14 & 0.14 & 105 & 1065\\
                                  & False & 7 & 5.83 & 0.33 & 0.29 & 0.27 & 522 & -\\
    \hline
    o3-mini & False & 27 & 22.50 & 0.55 & 0.47 & 0.49 & 195 & -\\
    o4-mini & False & \textbf{59} & 49.17 & 0.68 & 0.67 & 0.66 & 248 & -\\
    \hline
\end{tabular}
    \caption{Performance of LLMs on the query set (120 runs: 12 questions × 10 runs). "T" = tool use; "C" = correct answers; "A", "P", "R", "F1" = average Accuracy, Precision, Recall, F1; "FP" = false positives; "TC" = total tool calls. Reasoning models run without tools as a baseline. Best results for non-reasoning and reasoning models highlighted in boldface.}
    \label{tab:main_results_single}
\end{table}

\subsection{The Agent's Reasoning Tasks}
To evaluate the agent's reasoning capabilities, we designed a benchmark of 12 query templates, each targeting a distinct aspect of graph reasoning. These templates are grouped into three main categories, described below.

\subsubsection{Retrieval and Aggregation:}
This class includes direct lookups and summarization tasks. The \texttt{node\_by\_property\_query} asks the agent to find all nodes of a specific type with a property matching a given value, while \texttt{relationship\_by\_property\_query} requires finding all relationships of a certain type where a property matches a value, returning the connected nodes. Aggregation tasks such as \texttt{node\_count\_query} and \texttt{relationship\_count\_query} challenge the agent to count nodes or relationships matching certain criteria, and \texttt{node\_with\_most\_relationships} asks for the node with the highest number of outgoing relationships of a specific type.

\subsubsection{Path and Relational Traversal:}
These queries test the agent's ability to perform multi-hop reasoning and follow connections across the graph. The \texttt{path\_finding\_query} involves finding all paths connecting a source node type to a target node type through a specified intermediate node type. The \texttt{variable\_hop\_path\_query} requires the agent to find paths of variable length between two node types, followed by an additional step to any other node. The \texttt{path\_from\_specific\_node\_query} asks which nodes of a target type can be reached within a specific number of hops from a given starting node, and the \texttt{remote\_node\_property\_query} requires finding a node reachable in two or more hops (but not directly) and returning one of its properties.

\subsubsection{Complex Logical Composition.}
These queries mirror constructs from first-order logic, aiming to test the agent's ability to handle conjunctions and negation. The \texttt{compositional\_intersection\_query} requires identifying nodes that satisfy two independent relational conditions simultaneously, equivalent to a logical AND operation, $(\exists y \, R(x,y)) \land (\exists z \, S(x,z))$. Negation is tested with \texttt{negation\_with\_connection\_query}, which asks for nodes connected to a "positive" target type but not to a "negative" target type, corresponding to $(\exists y \, P(x,y)) \land \neg(\exists z \, Q(x,z))$, and with \texttt{negation\_on\_rel\_property\_query}, which requires finding nodes connected by a specific relationship type where a property on that relationship does not equal a certain value.

The exact templates and parameters for all 12 query types are detailed in appendix~\ref{appendix:question_details}.

\subsection{Experimental Setup}
We evaluated seven language models: gpt-4.1-nano, gpt-4.1-mini, gpt-4.1~\cite{openai2025gpt41}, gpt-4o-mini, gpt-4o~\cite{openai2024gpt4o}, o3-mini, and o4-mini~\cite{openai2025o3o4} (reasoning models at default effort) on 12 query templates across 10 independently generated graph instances per model, yielding 120 queries per model (12 templates $\times$ 10 runs). Each instance uses a distinct, different random graph with question templates filled by randomly sampled entities, and we ensure every question has a guaranteed answer. All labels, node types, relationship classes, and property values are random 4–8 character strings verified as non-words. Models are tested in two configurations: with tool access (schema + toolset only) and without tools (full graph in context). Prompt and schema details are given in Appendix~\ref{appendix:graph_schema_prompt}.
Primary experiments (Table~\ref{tab:main_results_single}) used graphs of up to 100 nodes, 4 node classes, 2 relationship classes, 3 properties per node and relationship on average, and 5 possible values per property. For scaling experiments (Table~\ref{tab:graph_size_effect}, Figure~\ref{fig:graph_size_performance}) we increased graph size to 150, 200, and 500 nodes, expanded to 8 node classes and 4 relationship classes, raised average properties to 6 per node and relationship, and doubled the value pool to 10 per property.

\subsection{Scope and Implications} 
It is important to note that while our benchmark of 12 query templates is diverse in complexity, it is not exhaustive of all possible graph reasoning tasks. However, these templates were designed to span \textit{fundamental graph operations} underlie most common use cases when users interact with knowledge graphs and they represent typical query patterns that more complex queries decompose into. Our results show that current LLMs struggle even on these basic operations (Table~\ref{tab:category_performance}), revealing that \textit{the challenge lies in fundamental graph reasoning rather than in the complexity of the query language.} 

\section{Discussion}
\subsection{The Baseline Evaluations}
Table~\ref{tab:main_results_single} lays out specific performance patterns across model configurations. Among non-reasoning models equipped with tools, gpt-4.1 achieves the highest performance, while gpt-4o-mini without tool access demonstrates the weakest results. As expected, o4-mini, a reasoning model, outperforms all other configurations, which can be attributed to both it being designed for multi-step reasoning and the fact that the 100-node graphs remain fully within its context window. The most noticeable pattern is the consistent performance gain when models use graph traversal tools with improvements in accuracy, precision, recall, and F1 scores, which shows that using graph-traversal tools provide more reliable reasoning. Tool-equipped models also show significantly lower false positive rates, even surpassing reasoning models. This suggests that deterministic tool execution reduces hallucination by enforcing factual grounding through verifiable graph operations. Another noteworthy observation is that gpt-4.1 with tool access outperforms o3-mini, which suggests that structured access to information through specialized tools can compensate for, and in some cases exceed implicit reasoning capabilities when the task involves verifiable operations over structured data.

\subsection{Performance by Question Category}
Table~\ref{tab:category_performance} provides a granular breakdown of model performance across twelve distinct query types. Models demonstrated the highest proficiency on single-step, direct retrieval tasks such as "Node by Property" queries. Similarly, models performed well on "Path from Specific Node", likely because the maximum path length was constrained to three and kept the search space manageable. The strong performance on "Negation on Rel Property" may be attributed to the detailed and explicit nature of this category's question templates, which aids models in both no-tools and with-tools settings. 
"Compositional Intersection" and "Negation with Connection" highlight the difficulty models face in applying multiple logical constraints (AND/NOT) simultaneously and are the most challenging categories. Aggregation-based queries like "Node Count" and "Relationship Count" also proved exceptionally difficult. The complete failure on Variable Hop Path queries is particularly telling, as these queries require the construction of lengthy answers that appear to exceed the models' capacity for complex query planning. These patterns are consistent across tool-equipped and no-tool settings, though tool access generally improves performance within each category.

{\footnotesize
\setlength{\tabcolsep}{0.5pt}
\renewcommand{\arraystretch}{1.0} 

\begin{longtable}{
     >{\raggedright\footnotesize\arraybackslash}p{3.8cm}|
    >{\centering\footnotesize\arraybackslash}p{0.7cm}|
    >{\centering\footnotesize\arraybackslash}p{0.45cm}
    >{\centering\footnotesize\arraybackslash}p{0.45cm}|
    >{\centering\footnotesize\arraybackslash}p{0.45cm}
    >{\centering\footnotesize\arraybackslash}p{0.45cm}|
    >{\centering\footnotesize\arraybackslash}p{0.45cm}
    >{\centering\footnotesize\arraybackslash}p{0.45cm}|
    >{\centering\footnotesize\arraybackslash}p{0.45cm}
    >{\centering\footnotesize\arraybackslash}p{0.45cm}|
    >{\centering\footnotesize\arraybackslash}p{0.45cm}
    >{\centering\footnotesize\arraybackslash}p{0.45cm}|
    >{\centering\footnotesize\arraybackslash}p{0.6cm}|
    >{\centering\footnotesize\arraybackslash}p{0.6cm}
}

\hline
\multirow{2}{*}{\textbf{Category}} & 
\multirow{2}{*}{\textbf{Total}} &
\multicolumn{2}{c|}{\textbf{gpt-4o-mini}} &
\multicolumn{2}{c|}{\textbf{gpt-4o}} &
\multicolumn{2}{c|}{\textbf{gpt-4.1-nano}} &
\multicolumn{2}{c|}{\textbf{gpt-4.1-mini}} &
\multicolumn{2}{c|}{\textbf{gpt-4.1}} &
\multicolumn{1}{c|}{\textbf{o3-mini}} &
\multicolumn{1}{c}{\textbf{o4-mini}} \\
\cline{3-14}
& & \textbf{T} & \textbf{NT} & \textbf{T} & \textbf{NT} & \textbf{T} & \textbf{NT} & \textbf{T} & \textbf{NT} & \textbf{T} & \textbf{NT} & \textbf{NT} & \textbf{NT} \\
\hline
\endfirsthead

\multicolumn{14}{c}%
{{\bfseries \tablename\ \thetable{} -- continued from previous page}} \\
\hline
\multirow{2}{*}{\textbf{Category}} & 
\multirow{2}{*}{\textbf{Total}} &
\multicolumn{2}{c|}{\textbf{GPT-4o-mini}} &
\multicolumn{2}{c|}{\textbf{GPT-4o}} &
\multicolumn{2}{c|}{\textbf{GPT-4.1}} &
\multicolumn{2}{c|}{\textbf{GPT-4.1-mini}} &
\multicolumn{2}{c|}{\textbf{GPT-4.1-nano}} &
\multicolumn{1}{c|}{\textbf{o3-mini}} &
\multicolumn{1}{c}{\textbf{o4-mini}} \\
\cline{3-14}
& & \textbf{T} & \textbf{NT} & \textbf{T} & \textbf{NT} & \textbf{T} & \textbf{NT} & \textbf{T} & \textbf{NT} & \textbf{T} & \textbf{NT} & \textbf{NT} & \textbf{NT} \\
\hline
\endhead

\hline
\endfoot
\endlastfoot

Node Count 
 & 1 & 0 & 0 & 0 & 0 & 0 & 1 & 0 & 0 & 0 & 0 & 0 & 0  \\[0.5ex]
\hline
Relationship Count  & 2 & 0 & 0 & 0 & 0 & 0 & 1 & 0 & 1 & 0 & 0 & 0 & 0  \\[0.5ex]
\hline
Node with Most Relationships  & 26 & 3 & 0 & 3 & 0 & 0 & 1 & 3 & 2 & 3 & 2 & 1 & 8  \\[0.5ex]
\hline
Node by Property &  85 & 10 & 0 & 10 & 6 & 7 & 0 & 10 & 5 & 10 & 7 & 10 & 10  \\[0.5ex]
\hline
Relationship by Property  & 9 & 0 & 0 & 0 & 0 & 0 & 0 & 1 & 0 & 0 & 0 & 0 & 8  \\[0.5ex]
\hline
Path Finding  & 7 & 0 & 0 & 0 & 0 & 0 & 0 & 0 & 0 & 1 & 1 & 0 & 5  \\[0.5ex]
\hline
Variable Hop Path  & 0 & 0 & 0 & 0 & 0 & 0 & 0 & 0 & 0 & 0 & 0 & 0 & 0 \\
\hline
Path from Specific Node  & 65 & 2 & 4 & 8 & 6 & 2 & 3 & 8 & 4 & 7 & 7 & 4 & 10 \\[0.5ex]
\hline
Remote Node Property  & 30 & 1 & 1 & 4 & 3 & 1 & 0 & 4 & 3 & 4 & 3 & 2 & 4 \\[0.5ex]
\hline
Compositional Intersection & 5 & 1 & 0 & 0 & 0 & 0 & 0 & 0 & 0 & 0 & 0 & 1 & 3 \\[0.5ex]
\hline
Negation with Connection & 3 & 0& 0& 0& 0& 0& 0& 0& 2& 0& 0& 0& 1 \\[0.5ex]
\hline
Negation on Rel Property & 55 &  1& 1& 9& 3& 1& 1& 6& 3& 10& 1& 9& 10 \\[0.5ex]
\hline
\caption{Performance by question category across both settings. "T"=tool use, "NT"= no-tools.} \label{tab:category_performance} 
\end{longtable}
}

\subsection{Experiments with Graph Size}

\begin{figure}[t]
\centering
\includegraphics[width=0.5\columnwidth]{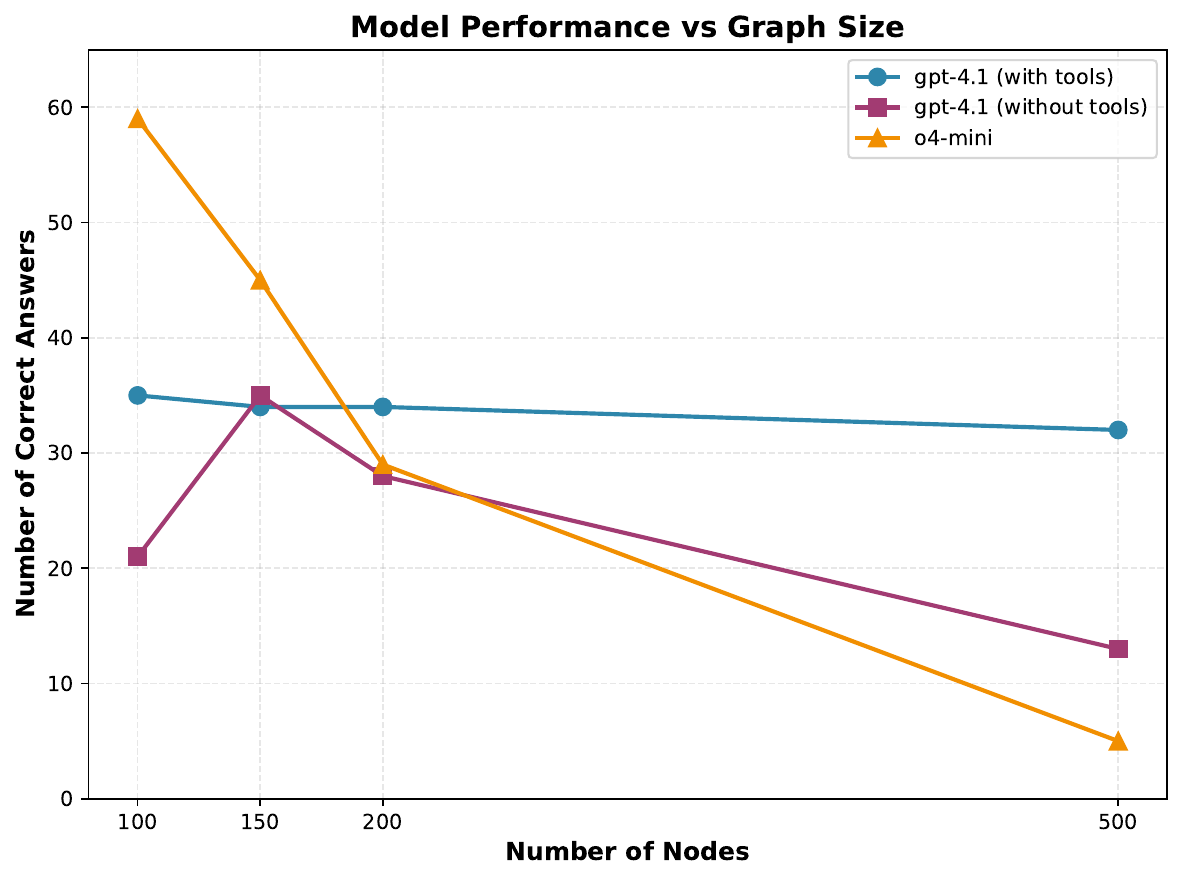}
\caption{Effect of graph size on model performance. The number of correct answers (out of 120 total) decreases as graph complexity increases for all models. Tool-equipped models demonstrate better robustness to increasing graph size compared to models without tool access.}
\label{fig:graph_size_performance}
\end{figure}

\begin{table}[t]
    \small
    \centering
    \vspace{0pt}
    \setlength{\tabcolsep}{6pt}
    \begin{tabular}{llcccc}
    \hline
    \textbf{Model} & \textbf{Metric} & \textbf{100} & \textbf{150} & \textbf{200} & \textbf{500} \\
    \hline
    \multirow{5}{*}{\shortstack{gpt-4.1\\(w.t)}} 
        & Acc. & 29.17 & 28.33 & 28.33 & \textbf{26.67} \\
        & P. & 0.47 & 0.39 & 0.40 & \textbf{0.39} \\
        & R. & 0.41 & 0.37 & 0.38 & \textbf{0.35} \\
        & F1 & 0.43 & 0.37 & 0.38 & \textbf{0.35} \\
        & F.P. & 232 & 149 & 170 & 151 \\
    \hline
    \multirow{5}{*}{\shortstack{gpt-4.1\\(w.o.t)}} 
        & Acc. & 17.5 & 29.17 & 23.33 & 10.83 \\
        & P. & 0.52 & 0.43 & 0.48 & 0.37 \\
        & R. & 0.49 & 0.49 & 0.50 & 0.33 \\
        & F1 & 0.48 & 0.43 & 0.46 & 0.31 \\
        & F.P. & 1006 & 1182 & 800 & 957 \\
    \hline
    \multirow{5}{*}{o4-mini} 
        & Acc. & 49.17 & 37.5 & 24.17 & 4.17 \\
        & P. & 0.68 & 0.56 & 0.40 & 0.13 \\
        & R. & 0.67 & 0.51 & 0.33 & 0.08 \\
        & F1 & 0.66 & 0.52 & 0.35 & 0.09 \\
        & F.P. & 248 & 108 & 44 & \textbf{108} \\
    \hline
\end{tabular}
    \caption{    Effect of graph size on model performance across node counts. "w.t" = with tools, "w.o.t" = without tools. Acc., P., R., F1, F.P. = average Accuracy, Precision, Recall, F1, and False Positives. Best results at 500 nodes in boldface.
    }
    \label{tab:graph_size_effect}
\end{table}

Table~\ref{tab:graph_size_effect} and Figure~\ref{fig:graph_size_performance} present results from our scalability experiments, in which we tested gpt-4.1 (the best-performing non-reasoning model) and o4-mini across increasing graph sizes. The findings reveal a critical advantage of tool-based navigation: gpt-4.1 with tools maintains relatively stable performance across all graph sizes. In contrast, both gpt-4.1 without tools and o4-mini demonstrate substantial performance decline as graph complexity increases. The degradation is particularly dramatic for o4-mini, which drops from 59 correct answers at 100 nodes to merely 5 at 500 nodes-despite the graph still fitting within its context window. This failure at scale demonstrates a fundamental limitation of in-context approaches, and exposes the inability of even reasoning-specialized models to maintain coherent graph traversal when processing entire structures inserted into the context window.

\subsection{Tool Minimalism as a Design Principle}

A central design choice in GraphWalk is the use of a minimal, orthogonal toolset. This stands in contrast to frameworks like KG-Agent \cite{jiang2025kg}, which employ rich toolboxes with 13 specialized tools organized into extraction tools, semantic tools, and logic tools. KG-Agent mainly evaluates if fine-tuned agents can solve KGQA when given curated, task-specific tools. While such specialization can improve task-specific performance, it conflates two distinct capabilities: the model's reasoning ability and the intelligence encoded in the tools. GraphWalk's minimal tools carry no domain intelligence, ensuring that all reasoning must originate from the LLM itself. This design choice has \textbf{generalizability} as a practical consequence. Since these tools are basic graph operations, they can apply unchanged to any graph-structured problem. We demonstrated this concretely by applying the same paradigm to both knowledge graph QA and maze traversal (Section~\ref{sec:maze}) without modifying the core framework.

\subsection{Model Limitations and Failure Modes}
Our evaluation revealed several recurring failure modes that illuminate the specific challenges LLMs face when reasoning over graph structures. Without tool access, models consistently demonstrated incomplete exploration patterns, often fixating on the first few relationships of a node and abandoning searches when answers were not immediately apparent, even when correct information existed within the same node's connections in the context. Additionally, models frequently failed to correctly ground their reasoning by selecting inappropriate starting nodes, leading to hallucinated entry points that derailed entire reasoning chains. Tool-based agents, while substantially more effective overall, exhibited their own characteristic failures. Across both tool settings and in both the experiments with maze traversal and synthetic graphs, we observed a persistent "last mile" problem: models would successfully gather correct evidence but fail to format responses according to the specified JSON schema, instead providing conversational answers that demonstrated correct reasoning but violated output requirements. These observations underscore both the value of structured tool access for graph navigation and the continued challenges in strategic re-planning and instruction adherence.

\section*{Conclusion}

In conclusion, our results demonstrate how GraphWalk can be used to substantially lift the performance of non-reasoning LLMs in both maze navigation as well  as graph Q\&A tasks and reach an equivalent performance level of much larger reasoning models. This is achieved by providing the models with a reasoning-harness through which they can break down the task into a series of traversal steps through a graph, i.e., the reasoning-harness allows the non-reasoning models to emulate reasoning chains in order to solve the task. Since many problems can be represented as graphs, we expect the applicability of GraphWalk to extend beyond the  specific domains tested in this work. The emulated reasoning chains further provide a level of transparency to the model’s decision-making process which otherwise would not be available. Similar to real reasoning models, GraphWalk  can also encounter failure modes such as getting stuck in a reasoning loop, e.g., a cycle in the graph, or finding it challenging to respond once the number of tool calls i.e., reasoning steps, gets too large. This  opens up interesting future directions for improving the framework, by incorporating for example a compaction mechanism for very-large tool-call chains.

\bibliographystyle{unsrtnat}
\bibliography{references}  

\appendix

\section{Appendix: Graph Schema and The Prompt}
\label{appendix:graph_schema_prompt}

In this section, we provide detailed information about our proposed framework. We present specifics on an example of a complete graph schema, details about the procedural generation of the graphs, a comprehensive description of the agent's tools, and the full templates for the experimental questions used in our benchmark.

\section{Synthetic Graph Generation}
\label{appendix:graph_generation}

Our synthetic graph generator produces random property graphs with configurable parameters. The generation process proceeds as follows:

\paragraph{Structure Generation.} Given parameters for the number of nodes $N$, node classes $C_n$, and relationship classes $C_r$, we first assign each node to a class uniformly at random. Edges are then generated: for each relationship class, we create directed edges between nodes of the appropriate source and target classes according to the schema, with edge density controlled by a configurable parameter.

\paragraph{Property Assignment.} Each node class and relationship class is assigned a set of property keys (with configurable average count). For each property key, a pool of possible values is generated; either random strings (4-8 characters, verified against an English dictionary to ensure non-semantic content) or random floating-point numbers. Each node or relationship is then assigned property values drawn uniformly from the corresponding pool.

\paragraph{Label Generation.} All entity names, including node class labels, relationship class names, and property key names are generated as random strings of characters with configurable length. Each generated string is checked against an English dictionary to ensure it carries no semantic meaning.

\paragraph{Question Instantiation.} For each graph instance, the 12 query templates are instantiated by selecting entities, properties, and values uniformly at random from the generated graph, with the constraint that each question has at least one valid answer. Ground truth is computed by executing the corresponding Cypher query against the Neo4j database \cite{neo4j}.

\paragraph{Output Format Handling and Evaluation.} A recurring challenge observed across models was non-compliance with the specified JSON output format. Models occasionally produced correct answers embedded within plain text rather than adhering to the structured format required for automated evaluation. To ensure comprehensive assessment of model capabilities, we implemented a two-stage extraction process. When standard JSON parsing failed, we employed an LLM-based extractor (powered by gpt-4o-mini) to extract the substantive answer from the model's output.

\subsection{An Example of the Graph Schema}

The graph schema is the only structural information provided to the LLM agent at the beginning of each task. It defines the blueprint of the graph, outlining all possible node types, relationship classes, and their associated properties. This information is critical, as the agent must use it to formulate valid traversal and query plans.

The schema is presented to the agent in a structured format that details each entity's type, name, Cypher representation, and available properties. As all names are randomly generated, the agent cannot infer any semantic meaning and must rely entirely on this structural definition.

Table \ref{tab:graph_schema_example} provides a representative example of the schema for one of the randomly generated graphs used in our experiments, which contained 100 nodes. This demonstrates the non-semantic nature of the labels and the structure the agent must interpret.

\begin{table}[h!]
\centering
\renewcommand{\arraystretch}{1.2} 
\begin{tabular}{c c c p{4cm} c}
\hline
\textbf{\#} & \textbf{Entity Type} & \textbf{Entity Name} & \textbf{Cypher Pattern} & \textbf{Property} \\
\hline
0 & Node & \texttt{Cevaz} & \texttt{(:Cevaz)} & \texttt{bexame} \\
1 & Node & \texttt{Cevaz} & \texttt{(:Cevaz)} & \texttt{key} \\
2 & Node & \texttt{Cevaz} & \texttt{(:Cevaz)} & \texttt{tanu} \\
... & ... & ... & ... & ... \\
\hline
18 & Relation-ship & \texttt{EPUQOSS} & \texttt{(:Cevaz)-[:EPUQOSS]->\newline(:Egodpw)} & \texttt{ukog} \\
19 & Relation-ship & \texttt{EPUQOSS} & \texttt{(:Cevaz)-[:EPUQOSS]->\newline(:Egodpw)} & \texttt{uqpc} \\
... & ... & ... & ... & ... \\
\hline
34 & Relation-ship & \texttt{LAJOZOS} & \texttt{(:Cevaz)-[:LAJOZOS]->\newline(:Cevaz)} & \texttt{bzle} \\
35 & Relation-ship & \texttt{LAJOZOS} & \texttt{(:Cevaz)-[:LAJOZOS]->\newline(:Cevaz)} & \texttt{uhiro} \\
... & ... & ... & ... & ... \\
\hline
\end{tabular}
\caption{A truncated example of the schema provided to the agent for a 100-node graph. The schema lists all node types, relationship types, and their respective properties.}
\label{tab:graph_schema_example}
\end{table}

\subsection{Prompt Structure for the With-Tools Setting}

The agent's reasoning process is initiated with a structured prompt designed for clarity and efficiency. The prompt consists of three core components: the graph schema and the natural language question. This minimalistic design compels the model to rely solely on the provided schema and tool definitions to construct a valid reasoning plan.

The exact template for the prompt provided to the agent is shown below. Placeholders like \texttt{\{graph\_schema\}} are dynamically populated at runtime.

\vspace{0.3cm}
\noindent\hrulefill

\begin{small}
\begin{verbatim}
You are a helpful assistant helping with Neo4j graph database in a controlled environment. 
At your disposal, you have a variety of tools, each specialized in performing a distinct
type of task. For successful task completion, based on the schema representation of the 
database, consider the task at hand and determine which tool or set of tools is best 
suited based on its capabilities and the nature of the query. Each one of the tools is 
equivalent to a cypher query. You can call the tools to query the graph database and 
extract the necessary information, but you cannot write a query yourself. Please note 
that in order to get the right answer, you might need to traverse the entire 
graph database.

This is the graph schema representation 
of the database:
<graph_schema>
{graph_schema}
</graph_schema>

System time:
<system_time>
{system_time}
</system_time>

<guidelines>
- Think step by step.
- If property values in the graph schema end with '...', it means the list is not 
    exhaustive and you should obtain the full list from the graph database if needed.
- Use the tools to query the graph database and extract the necessary information.
- Remember that unless otherwise specified the tools are DETERMINISTIC, which means 
calling them with the same arguments again will return the same result 
and should be avoided.
- Provide the answer in the correct format that is requested in the question.
- If the user query is not answered by the tools, ask for additional information.
- Continue calling tools until you have all the necessary information needed to answer 
the user query. When you have the final answer, STOP CALLING ANY TOOLS.
</guidelines>
\end{verbatim}
\end{small}

\noindent\hrulefill
\vspace{0.3cm}

\subsection{Prompt Structure for the No-Tools Setting}
The prompt used for No-Tools setting in our synthetic graph experiments in which the entire graph is injected into the prompt is given below.

\vspace{0.3cm}
\noindent\hrulefill

\begin{small}
\begin{verbatim}
You are a helpful assistant that answers questions about a graph database.
You are given the entire graph structure and its values in the format below.
Use this information to answer the user's question. 

The graph is represented below:
<graph_data>

{graph_data}

</graph_data>

System time:
<system_time>

{system_time}

</system_time>

Based on the graph data provided, please answer the following question.
variables:
- graph_data
- system_time 
\end{verbatim}
\end{small}

\noindent\hrulefill
\vspace{0.3cm}

\section{Agent Toolset Descriptions} \label{appendix:agent_tools}

The agent is equipped with a minimal, orthogonal set of four tools to interact with the graph database. Three of these tools (\texttt{get\_node\_by\_property}, \texttt{get\_all\_nearest\_neighbors}, \texttt{get\_unique\_property\_values}) are deterministic functions that map directly to Cypher queries. The fourth tool, \texttt{think}, is a special non-deterministic tool that allows the agent to record its reasoning process.

The exact docstrings provided to the agent for each tool are detailed below.

\subsection{\texttt{get\_node\_by\_property}}

\noindent\hrulefill

{\small\ttfamily\noindent
get\_node\_by\_property(label, property\_name, property\_value):\\
"""Retrieve a specific node from the graph database by matching a property value.

This tool searches for nodes with a specific label that have a particular property 
set to a given value. It's the primary way to find specific entities in the graph
when you know their identifying property (like name, ID, or other unique attribute).

Use this when you need to:
\begin{itemize}
\item Find a specific person, organization, drug, or other entity by name
\item Locate nodes with specific IDs or codes
\item Search for entities with particular attributes
\end{itemize}

Args:\\
\hspace*{1em}label (str): The node label/type (e.g., "Person", "Drug", "Company"). \\
\hspace*{3em}Must match exactly with labels in the graph schema.\\
\hspace*{1em}property\_name (str): The property to search by (e.g., "name", "id", "code").\\
\hspace*{3em}Must be a valid property for the specified label.\\
\hspace*{1em}property\_value: The exact value to match. Can be string, number, or other types\\
\hspace*{3em}depending on the property. Must match exactly (case-sensitive for strings).

Returns:\\
\hspace*{1em}list: List of matching nodes with all their properties. Each node is a dictionary\\
\hspace*{3em}containing all property key-value pairs for that node.

Example:\\
\hspace*{1em}get\_node\_by\_property("Person", "name", "John Smith")\\
\hspace*{1em}Returns: [\{"name": "John Smith", "age": 30, "id": "person\_123"\}]\\
"""}

\noindent\hrulefill

\subsection{\texttt{get\_all\_nearest\_neighbors}}

\noindent\hrulefill

{\small\ttfamily\noindent
get\_all\_nearest\_neighbors(label, property\_name, property\_value):\\
"""Get all directly connected neighbors of a specific node in the graph database.

This tool finds a node by its property value and returns ALL nodes that are directly
connected to it through any type of relationship. This is useful for exploring the
immediate neighborhood of a node and understanding its direct connections.

Use this when you need to:
\begin{itemize}
\item Explore what entities are directly connected to a specific node
\item Find all immediate relationships of a person, organization, or other entity
\item Discover the local neighborhood around a node
\item Get a comprehensive view of direct connections before drilling down
\end{itemize}

Args:\\
\hspace*{1em}label (str): The label/type of the central node (e.g., "Person", "Drug", "Company").\\
\hspace*{3em}Must match exactly with labels in the graph schema.\\
\hspace*{1em}property\_name (str): The property to identify the central node (e.g., "name", "id").\\
\hspace*{3em}Must be a valid property for the specified label.\\
\hspace*{1em}property\_value: The exact value to match for finding the central node.\\
\hspace*{3em}Must match exactly (case-sensitive for strings).

Returns:\\
\hspace*{1em}list: List of all neighboring nodes with their properties and relationship information.\\
\hspace*{3em}Each result includes both the neighbor node data and the relationship that\\
\hspace*{3em}connects it to the central node.

Example:\\
\hspace*{1em}get\_all\_nearest\_neighbors("Person", "name", "John Smith")\\
\hspace*{1em}Returns all people, organizations, locations, etc. directly connected to John Smith\\
"""}

\noindent\hrulefill

\subsection{\texttt{get\_unique\_property\_values}}

\noindent\hrulefill

{\small\ttfamily\noindent
get\_unique\_property\_values(property\_name, entity\_name, entity\_type):\\
"""Retrieve all unique values for a specific property across all nodes or relationships of a given type.

This tool is essential for data exploration and understanding what values exist
in the database. It helps you discover available options, validate data, and
understand the scope of information available for a particular entity type.

Use this when you need to:
\begin{itemize}
\item Explore what values are available for a property (e.g., all company names, or all relationship weights)
\item Validate if a specific value exists before searching
\item Get a complete list of options for categorical properties
\item Understand the data distribution and available entities
\item Find all possible values to choose from when building queries
\end{itemize}

Args:\\
\hspace*{1em}property\_name (str): The name of the property to get values for (e.g., "name",\\
\hspace*{3em}"category", "status"). Must be a valid property in the schema\\
\hspace*{3em}for the specified entity.\\
\hspace*{1em}entity\_name (str): The node label or relationship type to examine (e.g., "Person", "Drug",\\
\hspace*{3em}"Company", "INTERACTS\_WITH"). Must match exactly with labels/types in\\
\hspace*{3em}the graph schema.\\
\hspace*{1em}entity\_type (str): The type of the entity to examine, which can be 'node' or 'relationship'.

Returns:\\
\hspace*{1em}list: A list of dictionaries, each containing a unique value for the specified\\
\hspace*{3em}property. The structure is [\{"values": value1\}, \{"values": value2\}, ...].

Example for a node:\\
\hspace*{1em}get\_unique\_property\_values("name", "Company", "Node")\\
\hspace*{1em}Returns: [\{"values": "Pfizer"\}, \{"values": "Johnson \& Johnson"\}, \{"values": "Merck"\}]

Example for a relationship:\\
\hspace*{1em}get\_unique\_property\_values("year", "MET\_IN", "Relationship")\\
\hspace*{1em}Returns: [\{"values": "2020"\}, \{"values": "2021"\}]\\
"""}

\noindent\hrulefill

\subsection{\texttt{think}}

\noindent\hrulefill

{\small\ttfamily\noindent
think(thought):\\
"""Record and process reasoning steps during graph traversal and query planning.

This tool allows you to document your thought process, reasoning steps, and intermediate
conclusions while working through complex graph queries. It's particularly valuable
for multi-step problems where you need to plan your approach, track progress, or
explain your reasoning.

Use this when you need to:
\begin{itemize}
\item Break down complex queries into logical steps
\item Document your reasoning for choosing specific tools or approaches
\item Summarize findings from previous tool calls before proceeding
\item Explain why you're taking a particular path through the graph
\item Keep track of progress in multi-step graph traversals
\item Clarify your understanding of the problem before answering
\end{itemize}

Args:\\
\hspace*{1em}thought (str): Your reasoning, observation, or plan. Can include analysis of\\
\hspace*{3em}previous results, next steps to take, or explanations of your\\
\hspace*{3em}approach to solving the user's query.

Returns:\\
\hspace*{1em}str: The same thought string you provided, allowing you to record and\\
\hspace*{3em}reference your reasoning process.

Example:\\
\hspace*{1em}think("I found John Smith in the database. Now I need to find his company\\
\hspace*{3em}affiliations by looking at his neighbors, then find other employees\\
\hspace*{3em}of those companies.")\\
"""}

\noindent\hrulefill
\section{Maze Traversal Prompts}
We layout the prompts used for the two settings used (with tools and without tools) for our maze traversal experiments.

\subsection{Maze Traversal Prompt for the No-Tools Setting}
\vspace{0.3cm}
\noindent\hrulefill

\begin{small}
\begin{verbatim}
You are an AI agent solving a maze navigation problem. Your goal is to find 
the shortest path from the Start cell (S, key: {start_node_key}) to the Goal
cell (G, key: {end_node_key}) in a grid-based maze.


The maze is represented as a visual grid where each character/number 
represents a cell:

{wall_legend_line}

- Numbers (0, 1, 2, ...) represent open path cells, with each number being 
the cell's unique key

- 'S' represents the Start cell

- 'G' represents the Goal cell


The maze is displayed with borders to clearly show cell boundaries:

- '|' represents vertical borders between cells

- '+' and '-' represent horizontal borders between rows

- These border characters are for display only and are NOT part of the maze itself


Each traversable cell displays its unique key (0-indexed position). Cell keys 
increase left-to-right, top-to-bottom. For example, in a 10x10 maze, the top-left 
cell is 0, the cell to its right is 1, and so on. The first cell of the second row 
is 10.


The maze is a {maze_size} grid.

<maze_data>

{maze_data}

</maze_data>


<guidelines>

- Analyze the maze representation provided above.

- Find the shortest path from the Start cell (S) to the Goal cell (G).

- {wall_guideline}

- Each traversable cell displays its unique key (the number shown inside the cell). 
Use these keys to specify your path. Ignore border characters ('|', '+', '-') as 
they are for display only.

- Note: The Start cell (S) has key {start_node_key} and the Goal cell (G) has key 
{end_node_key}. They display 'S' and 'G' instead of their numeric keys in the maze 
for clarity.

- When you have found the complete path, return your answer as a JSON object with 
the complete sequence of cell keys that form the shortest path from Start to Goal.

- Return ONLY the JSON output in the format: {output_schema}

</guidelines>
variables:
- maze_data
- maze_size
- output_schema
- start_node_key
- end_node_key
- wall_legend_line
- wall_guideline
\end{verbatim}
\end{small}

\noindent\hrulefill
\vspace{0.3cm}

\subsection{Maze Traversal Prompt for the With-Tools Setting}

\vspace{0.3cm}
\noindent\hrulefill

\begin{small}
\begin{verbatim}
You are an AI agent solving a maze navigation problem. Your goal is to 
find the shortest path from the Start cell (key: {start_node_key}) to 
the End cell (key: {end_node_key}) in a grid-based maze.


The maze is represented as a graph database where cells are nodes connected 
by ADJACENT relationships. At your disposal, you have a variety of tools, 
each specialized in performing a distinct type of task.
Each tool is equivalent to a cypher query on the graph database.


The maze is a {maze_size} grid.

<graph_schema>

{graph_schema}

</graph_schema>

<guidelines>

- Think step by step.

- Use the tools to query the graph database and find the path from 
start to end.

- Remember that tools are DETERMINISTIC - calling them with the same 
arguments will return the same result.

{tool_history_note}

- The euclidian-distance property of each cell is the Euclidean 
distance from the End cell. The closer the cell is to the end node, 
the smaller the euclidian-distance.

- Cells that have not been marked yet have a mark_order value of -1. 

- Continue using tools until you have found the complete path. When 
you have the final answer, STOP CALLING TOOLS and return the complete 
sequence of cell keys that form the shortest path from start to end 
in a list. Remove redundant cells which belong to unnecessary 
traversal. A hint to identify redundant cells is that when you sort 
by marked order, you see a large different in cell keys between 
consecutive cells. Construct shortest path by removing redundant 
cells, starting from the source node.  

- Return ONLY the JSON output in the format: {output_schema}

</guidelines>
variables:
- graph_schema
- maze_size
- tool_history_note
- start_node_key
- end_node_key
- output_schema

\end{verbatim}
\end{small}

\noindent\hrulefill
\vspace{0.3cm}
\section{Maze Traversal Toolset}
In order to demonstrate the framework's mechanism on a visual example, we designed two tools to support the agent in incremental exploration. Consistent with GraphWalk's minimalist design philosophy, these tools provide only local observability and traversal history. The agent must choose these actions to traverse from the start cell to the goal.

\noindent\hrulefill

\subsection{\texttt{get\_possible\_next\_cells}}

\noindent\hrulefill

{\small\ttfamily\noindent
get\_next\_possible\_cells(node\_id):\\
""Get the neighboring cells that are traversable from a given node.
    
    This tool identifies all path cells directly adjacent (up, down, left, right)
    to the specified node. It will:
    1. Check if the given input node is a valid traversable cell (not a wall).
    2. Mark the given input node as visited in the maze.
    3. Return a list of adjacent cells that are paths (not walls).
    
    Use this tool to explore the maze by checking which directions are open from your given position.

Args:\\
\hspace*{1em}node\_id (str): The ID of the cell to explore.\\

Returns:\\
\hspace*{1em}str: A message listing the traversable neighboring cells, \\
\hspace*{3em}Returns an error if the input node is a wall or invalid.

"""}

\noindent\hrulefill

\subsection{\texttt{get\_connected\_path}}

\noindent\hrulefill

{\small\ttfamily\noindent
get\_connected\_path():\\
"""Find the shortest connected path through all cells visited so far.
    
    This tool automatically retrieves the complete list of cells you have visited
    (in the order they were marked) Tand then finds the shortest valid connected path
    from the first visited cell to the last visited cell using only those cells.
    Each step in the path must be to an adjacent cell (up, down, left, right —
    no diagonals or jumps).
    
    Use-cases for this tool:
    - Derive the final answer path after finishing your exploration
    - Check whether the cells you have visited so far form a valid connected route
      from start to your current position
    
Returns:\\
\hspace*{1em} str: The shortest valid connected path found within the visited cells, or an\\
\hspace*{3em}error message if no valid path can be formed from the visited cells.

Example:\\
\hspace*{3em}get\_connected\_path()
\hspace*{3em}Returns: "Shortest path from cell 0 to cell 42: ['0', '1', '2', ..., '42'] \\
"""}

\noindent\hrulefill

\section{Question Templates and Categories}\label{appendix:question_details}

Our benchmark is composed of 12 distinct query templates, designed to evaluate a range of reasoning skills from simple retrieval to complex logical composition. The details for each template are provided in Table \ref{tab:query_templates}. For each template, we define the natural language instruction given to the agent, the ground-truth Cypher query that corresponds to the task, and the expected JSON output schema. Placeholders like \texttt{\{source\_label\}} are populated dynamically from the specific random graph being used for the test run.

{
\setlength{\tabcolsep}{2pt}

\begin{longtable}{
    >{\raggedright\small}p{1.4cm}
    >{\raggedright\scriptsize}p{1.5cm}          
    >{\raggedright\small}p{4.0cm}          
    >{\raggedright\small}p{2.9cm}
    >{\raggedright\arraybackslash\scriptsize}p{7.5cm}
}
\caption{Detailed Breakdown of the Query Templates} \label{tab:query_templates} \\
\hline\hline
\textbf{Category} & \textbf{Question Class} & \textbf{Instruction Template} & \textbf{Output Schema} & \textbf{Ground Truth (Cypher)} \\
\hline
\endfirsthead

\multicolumn{5}{c}%
{{\bfseries \tablename\ \thetable{} -- continued from previous page}} \\
\hline\hline
\textbf{Category} & \textbf{Question Class} & \textbf{Instruction Template} & \textbf{Output Schema} & \textbf{Ground Truth (Cypher)} \\
\hline
\endhead

\hline\hline
\endfoot

\multirow{8}{=}{{Retrieval \newline \& Aggre-gation}} & {Node Count} &
{Count the number of "\{source\_label\}" nodes
that are connected to any "\{target\_label\}" node. 
Return ONLY the output with the count in JSON format: \{\{output\_schema\}\}.} & 
\begin{minipage}[t]{\linewidth}\ttfamily
[\{"count": "number"\}]
\end{minipage} & 
\begin{minipage}[t]{\linewidth}\ttfamily
MATCH (a:\{source\_label\})
-->
(b:\{target\_label\}) 
RETURN count(DISTINCT a) AS count
\end{minipage} \\
\cline{2-5}

& {Relation-ship Count} &
{How many relationships of type "\{rel\_type\_name\}" exist? Return ONLY the output with the count in JSON format: \{\{output\_schema\}\}.} &
\begin{minipage}[t]{\linewidth}\ttfamily
[\{"count": "number"\}]
\end{minipage} &
\begin{minipage}[t]{\linewidth}\ttfamily
MATCH ()-[r:\{rel\_type\_name\}]->() 
RETURN count(DISTINCT r) AS count
\end{minipage} \\
\cline{2-5}

& {Node with Most Relationships} &
{Which "\{source\_node\_label\}" node has the most outgoing "\{rel\_type\_name\}" relationships? Return ONLY ONE answer in JSON format as per the schema: \{\{output\_schema\}\}.} &
\begin{minipage}[t]{\linewidth}\ttfamily
[\{"node\_key": "string", "rel\_count": "number"\}]
\end{minipage} &
\begin{minipage}[t]{\linewidth}\ttfamily
MATCH (n:\{source\_node\_label\})
      -[r:\{rel\_type\_name\}]
      ->() WITH n, count(r) AS rel\_count 
RETURN n.key AS node\_key,
rel\_count ORDER BY rel\_count 
DESC LIMIT 1
\end{minipage} \\
\cline{2-5}

& {Node by Property} &
{Find all "\{node\_label\}" nodes where "\{prop\_name\}" is "\{prop\_value\}". Return results in JSON format according to the schema: \{\{output\_schema\}\}.} &
\begin{minipage}[t]{\linewidth}\ttfamily
[\{"node\_key": "string"\}]
\end{minipage} &
\begin{minipage}[t]{\linewidth}\ttfamily
MATCH (n:\{node\_label\} \{ 
  \{prop\_name\}: \{query\_prop\_value\} 
\}) 
RETURN DISTINCT n.key AS node\_key
\end{minipage} \\
\cline{2-5}

& {Relation-ship by Property} &
{Find all "\{rel\_type\_name\}" relationships where "\{prop\_name\}" is "\{prop\_value\}". Return results in JSON format based on the schema: \{\{output\_schema\}\}.} &
\begin{minipage}[t]{\linewidth}\ttfamily
[\{"source\_key": "string", "target\_key": "string", ...\}]
\end{minipage} &
\begin{minipage}[t]{\linewidth}\ttfamily
MATCH (s)-[r:\{rel\_type\_name\} \{ 
  \{prop\_name\}: \{query\_prop\_value\} 
\}]->(t) 
RETURN s.key as source\_key, 
       t.key as target\_key, 
       \{return\_clause\}
\end{minipage} \\
\hline

\multirow{6}{=}{{Path \& Relational Traversal}} & {Path Finding} &
{Find all paths from "\{source\_label\}" to "\{target\_label\}" through "\{middle\_label\}". Return results in JSON format as per schema: \{\{output\_schema\}\}.} &
\begin{minipage}[t]{\linewidth}\ttfamily
[\{"source\_node\newline \_key": "string", "target\_node\_key": "string"\}]
\end{minipage} &
\begin{minipage}[t]{\linewidth}\ttfamily
MATCH (a:\{source\_label\})-->
      (b:\{middle\_label\})-->
      (c:\{target\_label\}) 
RETURN a.key AS source\_node\_key, 
       c.key AS target\_node\_key
\end{minipage} \\
\cline{2-5}

& {Variable Hop Path} &
{Find all paths where a "\{source\_label\}" node reaches a "\{target\_label\}" node in 1 to \{n\} steps, then takes one more step to any other node. Return the keys of the source and target nodes in JSON format as per schema: \{\{output\_schema\}\}.}&
\begin{minipage}[t]{\linewidth}\ttfamily
[\{"source\_node\newline\_key": "string", "target\_node\_key": "string"\}]
\end{minipage} &
\begin{minipage}[t]{\linewidth}\ttfamily
MATCH (a:\{source\_label\})
      -[*1..\{n\}]->\newline(b:\{target\_label\})
      -->() RETURN DISTINCT a.key 
      AS source\_node\_key, 
       b.key AS target\_node\_key
\end{minipage} \\
\cline{2-5}

& {Path from Specific Node} &
{Find all paths of 1 to \{n\} steps from the node with key "\{source\_key\}" to any node of type "\{target\_label\}". Return the keys of the target nodes found in JSON format: \{\{output\_schema\}\}.} &
\begin{minipage}[t]{\linewidth}\ttfamily
[\{"target\_node\newline\_key": "string"\}]
\end{minipage} &
\begin{minipage}[t]{\linewidth}\ttfamily
MATCH (a:\{source\_label\} 
\{key: '\{source\_key\}'\})
      -[*1..\{n\}]->
      (b:\{target\_label\}) 
RETURN DISTINCT b.key
AS target\_node\_key
\end{minipage} \\
\cline{2-5}

& {Remote Node Property} &
{From a "\{source\_label\}" node with key "\{source\_key\}" find a "\{target\_label\}" node that is not a direct neighbor but is reachable in 2 or more hops, and return its "\{prop\_name\}". ANY valid node's property will be accepted. Return ONLY ONE answer in JSON format: \{\{output\_schema\}\}.} &
\begin{minipage}[t]{\linewidth}\ttfamily
[\{"value": "\{prop\_type\}"\}]
\end{minipage} &
\begin{minipage}[t]{\linewidth}\ttfamily
MATCH (a:\{source\_label\} 
\{key: '\{source\_key\}'\})
      -[*2..\{self.max\_hops\}]
      ->(b:\{target\_label\}) 
WHERE NOT (a)-->(b) 
RETURN DISTINCT b.\{prop\_name\} as value
\end{minipage} \\
\hline

\multirow{4}{=}{{Complex Logical Compo-sition}} & {Compo-sitional Intersection} &
{Find all nodes of type "\{source\_label\}" that have a relationship to at least one "\{target1\_label\}" node AND at least one "\{target2\_label\}" node. Return the keys of these "\{source\_label\}" nodes in JSON in this format: \{\{output\_schema\}\}.} &
\begin{minipage}[t]{\linewidth}\ttfamily
[\{"node\_key": "string"\}]
\end{minipage} &
\begin{minipage}[t]{\linewidth}\ttfamily
MATCH (a:\{source\_label\}) 
WHERE EXISTS((a)
-->(:\{target1\_label\})) 
  AND EXISTS((a)
  -->(:\{target2\_label\})) 
RETURN DISTINCT a.key AS node\_key
\end{minipage} \\
\cline{2-5}

& {Negation with Connection} &
{Find all nodes of type "\{source\_label\}" that are connected to at least one "\{positive\_target\_label\}" node AND are not connected to any "\{negative\_target\_label\}" node. Return their keys in JSON in this format: \{\{output\_schema\}\}.} &
\begin{minipage}[t]{\linewidth}\ttfamily
[\{"node\_key": "string"\}]
\end{minipage} &
\begin{minipage}[t]{\linewidth}\ttfamily
MATCH (a:\{source\_label\}) 
WHERE EXISTS((a)
-->(:\{positive\_target\_label\})) 
  AND NOT EXISTS((a)
  -->
  
  (:\{negative\_target\_label\})) 
RETURN DISTINCT a.key AS node\_key
\end{minipage} \\
\cline{2-5}

& {Negation on Rel Property} &
{Find all "\{source\_label\}" nodes where "\{source\_prop\_name\}" is "\{source\_prop\_value\}". From those, find the ones connected to a "\{target\_label\}" node by a "\{rel\_type\_name\}" relationship where the relationship's "\{prop\_name\}" is not "\{val2\}". Return the keys of the source nodes in JSON in this format: \{\{output\_schema\}\}.} &
\begin{minipage}[t]{\linewidth}\ttfamily
[\{"node\_key": "string"\}]
\end{minipage} &
\begin{minipage}[t]{\linewidth}\ttfamily
MATCH (a:\{source\_label\} 

\{ \{source\_prop\_name\}: 

  \{query\_source\_prop\_value\} 
\})
-[r:\{rel\_type\_name\}]->

(b:\{target\_label\}) 
WHERE r.\{prop\_name\} 
<> \{query\_rel\_prop\_value\} 
RETURN DISTINCT a.key AS node\_key
\end{minipage} \\

\end{longtable}
}

\clearpage

\end{document}